\documentclass[10pt,letterpaper]{article}

\usepackage{cogsci}

\cogscifinalcopy 

\usepackage[
  style=apa,
  natbib=true,
  annotation=false,
]{biblatex}

\usepackage{float} 

\usepackage{graphicx} 

\usepackage[htt]{hyphenat}
\usepackage{xcolor}

\renewcommand{\thefootnote}{\fnsymbol{footnote}}

\title{Online library learning in human visual puzzle solving}

\author[1]{\mbox{Pinzhe Zhao (s2471381@ed.ac.uk)}}
\author[2,3]{\mbox{Emanuele Sansone}}
\author[4]{\mbox{Marta Kryven\thanks{These authors contribute equally.}}}
\author[1]{\mbox{Bonan Zhao\textsuperscript{*}}}
\affil[1]{University of Edinburgh}
\affil[2]{MIT CSAIL}
\affil[3]{KU Leuven}
\affil[4]{Dalhousie University}

\begin{document}

\maketitle

\renewcommand{\thefootnote}{\arabic{footnote}}
\setcounter{footnote}{0}

\begin{abstract}
When learning a novel complex 
task, 
people often form 
efficient reusable abstractions that simplify future work, despite uncertainty about the future. 
We study this process in a visual puzzle task where participants define and reuse helpers---intermediate constructions that capture repeating structure. In an online experiment, participants solved puzzles of increasing difficulty. 
Early on, they created many helpers, favoring completeness over efficiency. With experience, helper use became more selective and efficient, reflecting sensitivity to reuse and cost. Access to helpers enabled participants to solve puzzles that were otherwise difficult or impossible. Computational modeling shows that human decision times and number of operations used to complete a puzzle increase with search space estimated by a program induction model with library learning.
In contrast, raw program length predicts failure but not effort. Together, these results point to online library learning as a core mechanism in human problem solving, allowing people to flexibly build, refine, and reuse abstractions as task demands grow.

\end{abstract}


\section{Introduction}


From knitting a sweater to building a cathedral, people often tackle complex tasks by creating and manipulating reusable structural units. 
While these units---intermediate representations or procedures---help manage cognitive load and enable the otherwise intractable solutions \citep{gobet2001chunking,zhao2024model,gentner1983structure}, the decision of which units to build is surprisingly understudied \citep{ham2025teaching}. 
Future demands are often uncertain, and constructing a new unit may incur costs in time, memory, and effort \citep{wu2023chunking,o2009fragment}. 
If capturing the distribution of these units can be understood as a form of library learning \citep{ellis2023dreamcoder}, then a central challenge to cognition is library learning but online:
how do people build reusable abstractions on the fly?

Studying online library learning poses several methodological challenges. First, the space of possible helpers is often large and open-ended, making it difficult to characterize meaningful abstractions \citep{rule2024symbolic}. Second, helper creation trades off expressivity against efficiency: too many helpers streamline individual solutions while increasing the cost of maintaining and selecting among them \citep{wu2023chunking}. Third, behavioral tasks must be rich enough to elicit abstraction, yet structured enough to allow systematic analysis of learning trajectories.

In this paper, we address these challenges using a novel visual puzzle paradigm. In this paradigm, participants can build and reuse geometric patterns while solving progressively more complex tasks, allowing 
us to study abstraction creation and reuse on the fly.
We first motivate the task and its computational relevance, and then report an exploratory behavioral study examining how abstraction 
evolves over time.
To foreshadow, 
we found that human cognitive computations scale with the size of the underlying search space of a program induction model with online library learning, but can not be predicted by the raw number of primitives encoding a given puzzle, suggesting that human problem solving relies on online learning of adaptive structural abstractions. 

\section{Background}

Abstraction in problem solving has been studied extensively across multiple computational frameworks \citep[e.g.][]{gobet2001chunking,gentner1983structure,simon2013problem}. Here we focus on program induction approaches applied to visual puzzles.

\subsection{Problem solving as program induction}

A common computational perspective on puzzle-solving is to formalize this problem as program induction -- the process of inferring a program that generates the target solution from sparse input–output examples \citep{wang2023hypothesis,bramley2018learning,johnson2021fast}. 
These programs specify generative compositional structures,  
capturing key properties of human cognition such as abstraction, hierarchy, and relational reasoning \citep{gentner1983structure,goodman2008rational}, across domains such as compositional reasoning \citep{johnson2021fast}, few-shot learning 
\citep{lake2015human,cogsci2025,zhao2022people}, and recently, geometric puzzles \citep{he2025bootstrapping}.


Despite their broad successes, program induction models depend critically on choice of representations and priors, raising the question of how representations themselves are learned, reused, and adapted across tasks. Human chess players, for example, rely on effective representations that organize experience into chunks to substantially reduce search \citep{gobet2001chunking}. By contrast, many existing program induction models assume fixed hand-specified representations~\citep{lake2015human,franken2021algorithms}, limiting their ability to scale as search spaces grow exponentially with task complexity. 


\subsection{Library learning}

In response to these challenges, library learning in program induction models addresses how learners incrementally acquire reusable abstractions that support efficient problem solving across tasks \citep{ellis2023dreamcoder}. At the computational level of analysis \citep{marr1982vision}, inducing a library that captures recurring structure in the environment enables compact and generalizable representations. At the algorithmic level, libraries compress future search by biasing it toward previously useful subroutines,  reshaping the hypothesis space over time \citep{zhao2024model}. This perspective has been used to model abstraction \citep{rule2024symbolic}, skill acquisition \citep{tian2020learning}, and curriculum effects \citep{he2025bootstrapping}, providing a formal account of how experience across tasks can accumulate into structured knowledge rather than isolated solutions.

However, most existing library learning models rely on offline optimization, repeatedly re-solving all tasks to evaluate candidate libraries \citep{ellis2023dreamcoder,bowers2023top,cao2023babble}. While such procedures are well suited for finding the computational-level solutions, they contrast sharply with the demands of everyday cognition: people rarely encounter identical tasks multiple times, and must operate under uncertainty about future tasks, making global optimization infeasible. Instead, individual cognition must address the library learning problem \textit{online}, as experience unfolds.


\subsection{Visual puzzles}

To investigate online learning of complex structured representations, we adopt an experimental paradigm of a visual puzzle task.
Visual puzzles provide a rich testbed for studying human problem solving because they combine perceptual reasoning, planning, and abstraction. Prior work has shown that such tasks reveal how people identify patterns, sequence actions, and transfer knowledge across related problems~\citep{chu2025makes,he2025bootstrapping}. They are especially useful for studying stepwise reasoning and the emergence of intermediate strategies, as they make the process-level solution observable and tractable for behavioral analysis \citep{correa2025exploring}.

\begin{figure*}[t]
\centering
\includegraphics[width=.9\textwidth]{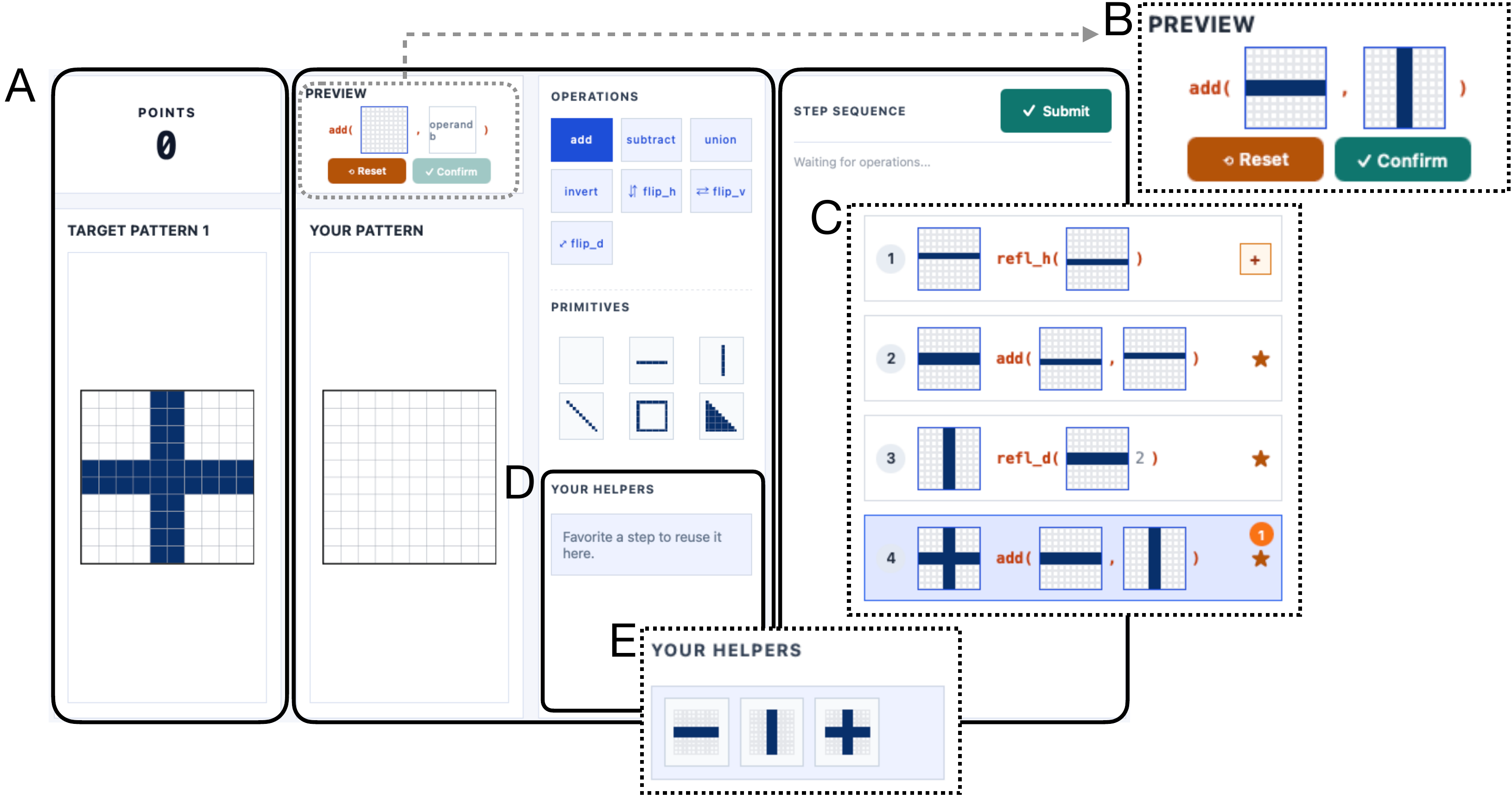}
\caption{Task interface. 
A. Starting view of a trial. Left: Total points so far and the target shape for this trial. 
Middle: Work space, matching the target shape with the provided Operations and Primitive shapes. 
Right: List of steps, each step corresponds to one line of program.
Black borders are only added in the paper for illustration.
B. Example preview.
C. Example programs. Each line has a line number, a thumbnail of the pattern that line creates, and the corresponding program (operations over primitives, steps, or helpers).
D. The initial helper space.
E. Helper space with example helpers.}
\label{fig:task_and_stimuli}
\end{figure*}

\section{Behavioral Paradigm: Pattern Builder}

We create a novel Pattern Builder Task (PBT) where people create progressively more complex visual patterns with access to creating reusable helpers that carry over to future tasks.



\subsection{Building patterns by composing programs}

In the PBT, participants see a target pattern on a $10 \times 10$ grid 
on the left side of the screen, and an empty canvas on the right (Figure~\ref{fig:task_and_stimuli}A). 
The goal is to 
recreate the target using a small set of geometric and transformation primitives, including five shapes: \texttt{line\_horizontal}, \texttt{line\_vertical}, \texttt{diagonal}, \texttt{square} (border only), \texttt{triangle};
three binary transformations: \texttt{add}, \texttt{subtract}, \texttt{overlap};
and four unary operations: \texttt{invert}, \texttt{reflect\_horizontal}, \texttt{reflect\_vertical}, \texttt{reflect\_diag}.
%
Participants can combine operations and primitives to progressively build a pattern, shown in the `Your Pattern' grid (Figure~\ref{fig:task_and_stimuli}B). 
In addition to the initial primitives, participants can also use any line from 
the `Step Sequence' as operants 
(Figure~\ref{fig:task_and_stimuli}C).

For example, to build a `thick cross,' we may first reflect a primitive horizontal line horizontally: \texttt{refl\_h(line\_h)} and combine this with another primitive horizontal line: 
\texttt{add(line\_h,refl\_h(line\_h))}, leading to a thick horizontal line.
Next, we reflect this thick line along the diagonal to create a thick vertical line: \texttt{refl\_d(add(line\_h,refl\_h(line\_h)))}, and add the two thick lines into the final cross: 
\texttt{add(add(line\_h,refl\_h(line\_h)),refl\_d(add(line\_h,refl\_h(line\_h))))}.
We encourage the reader to try the task at: \url{https://bococo-81.inf.ed.ac.uk/}.

\subsection{Creating and using helpers}

In PBT, participants can save any intermediate step as a \textit{helper} by clicking a `+' button. Helpers appear as thumbnails in `Your Helpers'  (Figure~\ref{fig:task_and_stimuli}D-E), and can be subsequently used as operands, effectively extending the primitive vocabulary. 
Saved helpers persist across trials, unless removed from the collection by clicking the `-' button.
For instance, after constructing an X-shaped pattern in Trial 5, a participant could save it as a helper and reuse it in Trials 6, 7, 13 and 14 which involve an X-shape structure, rather than rebuilding the X shape from primitives. 
By allowing participants to externalize and retain useful structures for reuse, this mechanisms allows us to observe their online library learning.



\subsection{Free play}

PBT also supports a free-play mode, 
where participants start with a fresh canvas, an empty helper library, no target to match, and can explore building any patterns they wish. They can submit their creations to a gallery,  giving them optional names. After a submission, the canvas and operation history are cleared, but helpers created persist through the entire free play phase. 
The free-play mode allows us to observe how people explore the compositional space they've learned in the absence of an instrumental task. 

\section{Computational Models}\label{sec:model}

We formalize the PBT problem by inductive program synthesis within the programming-by-example (PBE) paradigm. In PBE, a domain-specific language (DSL) defines the program search space, while a set of input–output examples constrains the space of valid solutions and captures user intention. In our setting, 
the DSL consists of the primitives and transformations as defined in the PBT. The goal of PBE induction is therefore to find a program $p$ in the given DSL such that 
executing $p$ on an empty canvas $c_0$ produces the target pattern $c^*$, i.e, $\llbracket p \rrbracket (c_0) = c^*$.

\subsection{Bottom-up search over primitives}

One way to solve the PBE induction is via exhaustive search. Specifically, each candidate program $p$ is represented as an abstract syntax tree, where leaves correspond to primitives in the DSL and internal nodes correspond to transformation operators (denoted as $\mathcal{T}$). We perform a bottom-up search \citep{albarghouthi2013recursive,udupa2013transit} 
over programs ordered by size. Let $|p|$ denote program size and let $\mathcal{H}_k$ be the set of programs of size $k$, with $\mathcal{H}_0$ consisting of only the primitives. Programs of size $k{+}1$ are generated by applying transformation operators to programs in $\mathcal{H}_{\le k}$:
\[
\mathcal{H}_{k+1} = \{\, t(p_1,\ldots,p_n) \mid t \in \mathcal{T},\; p_i \in \mathcal{H}_{\le k} \,\}.
\]
Search proceeds by enumerating $\mathcal{H}_0, \mathcal{H}_1, \mathcal{H}_2, \ldots$, corresponding to a breadth-first traversal of the program space.

To mitigate the combinatorial explosion inherent in exhaustive search, we prune the search space using the principle of observational equivalence \citep{albarghouthi2013recursive,udupa2013transit}. Concretely, we eliminate programs that are equivalent with respect to their output behavior, i.e., producing identical patterns when executed on an empty canvas. Observationally equivalent programs are ranked according to one of two criteria: lexicographic order (referred to as \textit{Baseline} in our experiments) or program length (referred to as \textit{Short}). Pruning is then performed by retaining only the top-ranked program within each equivalence class.

\subsection{Online library learning}

The bottom-up search algorithm described above 
guarantees completeness under \textit{Baseline} and minimality under \textit{Short}. However, it suffers from poor scalability in online settings, because the search must be restarted from the primitives for each new target pattern. 
To address these limitations, we introduce a simple form of library learning.
Let $\mathcal{P}$ denote the set of primitives in the DSL and let $p^{(i)}$ be the program discovered for pattern $i$. After solving pattern $i$, we augment the DSL by adding $p^{(i)}$ to the primitive set, i.e.,
\[
\mathcal{P}^{(i+1)} = \mathcal{P}^{(i)} \cup \{p^{(i)}\}.
\]
Subsequent searches are then performed over the expanded DSL, allowing previously discovered programs to be reused as atomic components. This mechanism reduces redundant computation and enables the construction of deeper programs within a fixed search budget \citep{zhao2024model}. We augment the two existing variants, \textit{Baseline} and \textit{Short}, with this library component, and refer to the resulting methods as \textit{Library} and \textit{Short+Library}, respectively.

\subsection{Human-model comparison}

Our models define two computational metrics: 
\textit{program length} and \textit{nodes expanded}.
Program length is the number of primitives in the shortest solution program found by \textit{Short+Library}, 
and served as a proxy for pattern difficulty.
\textit{Nodes expanded} counts each candidate program $p \in \mathcal{H}_{\le k}$ that the model evaluates during search, including programs considered before pruning via observational equivalence, reflecting the effective size of the explored program space.
We relate these computational metrics to human behavioral metrics: \textit{steps} taken to complete a pattern---the length of a human mental program compressed via helper reuse; \textit{solution time} in seconds---the amount of cognitive computations; \textit{accuracy}---defined as the exact match between a target patten and the produced output.
We also track \textit{success rate} for each pattern as a measure of population-level difficulty. 

\section{Behavioral Study}

We report an online behavioral experiment using the PBT, evaluate model predictions, and examine how helper creation and helper use evolve in human participants. 

\subsection{Experiment}
The experiment was approved by the Ethics Committee (ref. no. omitted for anonymity). 
All participants gave informed consent before participating.
Pre-registration is available at \url{https://aspredicted.org/ju2bw5.pdf}.

\paragraph{Participants}

Thirty-four participants were recruited from Prolific Academic and compensated at a rate of \pounds7.27/hr. Four participants were excluded based on pre-registered criteria for task disengagement (
rapid declining of performance and 
time investment in later trials), 
yielding a final sample of 30 participants (47\% female; $M_{\text{age}} = 36.2$, $SD = 9.8$). 
The full experiment took on average 50.7 minutes ($SD = 17.9$).  

\begin{figure}
    \centering
    \includegraphics[width=0.9\linewidth]{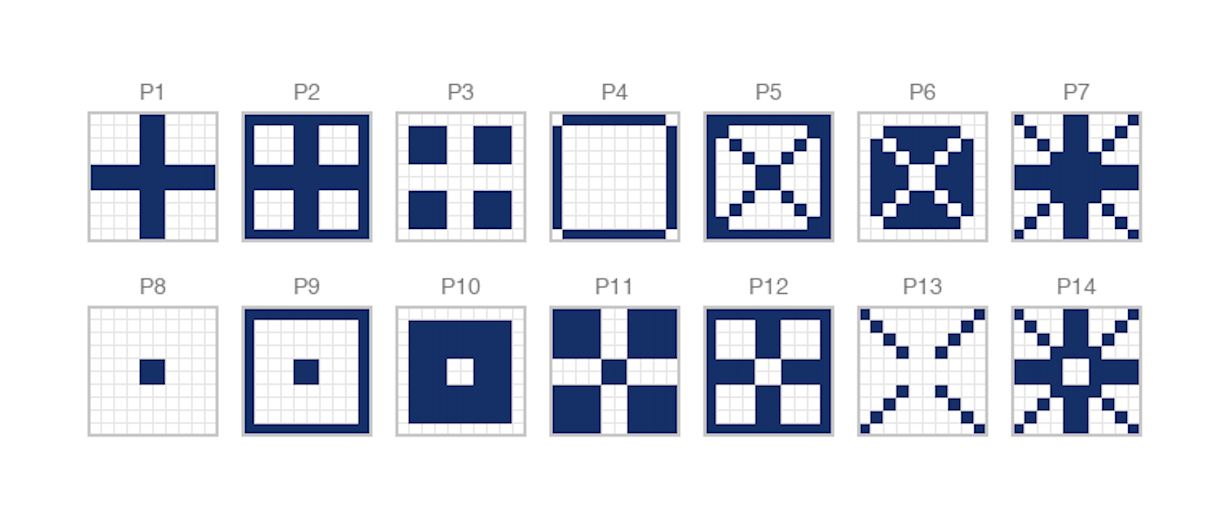}
    \caption{Target patterns used in the Experiment.}
    \label{fig:stimuli}
\end{figure}

\paragraph{Design}
The study was an exploratory, single-condition experiment.
Participants solved 14 target patterns using the PBT (Figure~\ref{fig:task_and_stimuli}),
 shown in Figure~\ref{fig:stimuli}, always presented 
 in the given order. 
The patterns gradually increased in difficulty, 
motivating increasing use of helpers. 
We refer to these patterns as \textit{P1 ... P14
}, in the order they appeared.  

\paragraph{Procedure}

After providing consent, 
participants read instructions describing the task and the interface. They then completed a PBT tutorial, 
followed by a brief comprehension quiz. 
Participants then proceeded to solve the 14 target patterns (Figure~\ref{fig:stimuli}), 
followed by a free play phase, with a minimum duration of 5 minutes.
Participants could continue beyond this limit with no time restriction, if they wished. 
The experiment concluded with a debrief.

\subsection{Results}


\paragraph{Participants and Library models converged on helper use.}

\begin{figure}[t]
  \centering
  \includegraphics[width=\columnwidth]{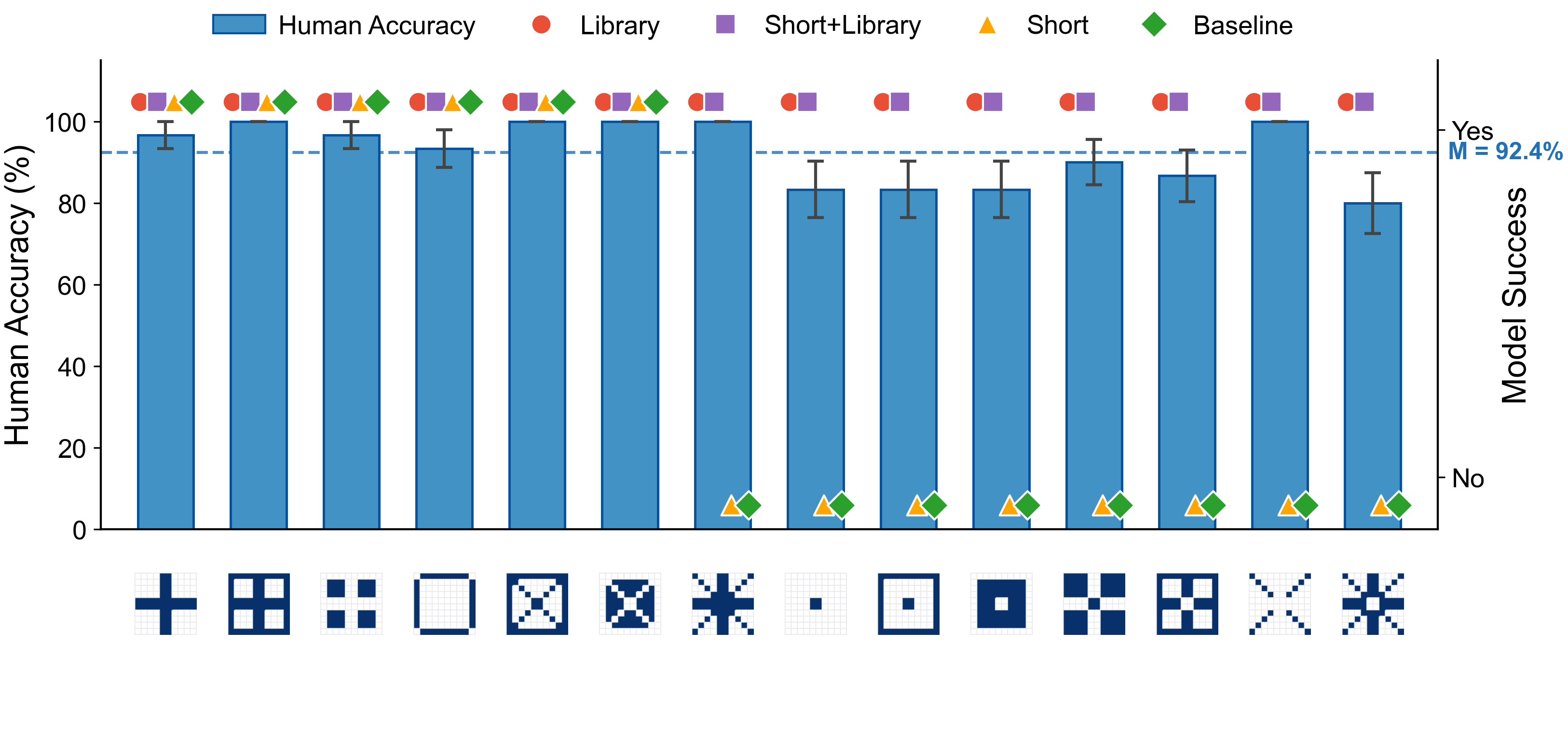}
  \caption{Accuracy 
  across 14 target patterns. Bars and standard errors for human participants. Colored dots for models. 
  }
  \label{fig:fig3}
\end{figure}

Participants and Library models were able to find solutions for all 14 puzzles by reusing past solutions. In contrast, Baseline models 
could only solve the first six puzzles within the given computational budget (Figure~\ref{fig:fig3}).
Over the 420 trials (30 participants $\times$ 14 patterns), participants achieved 
an overall accuracy of 92.4\% ($SD = 14.3\%$, 95\% CI $[87.3\%, 97.5\%]$). 
Across all successful trials, participants created a total of 470 helpers ($M = 15.7$, $SD = 9.0$). Most participants (93.3\%, 28/30) created at least one helper, with individual totals ranging from 0 to 35.
%

\begin{figure}[t]
\includegraphics[width=\columnwidth, height=0.3\textheight,keepaspectratio]{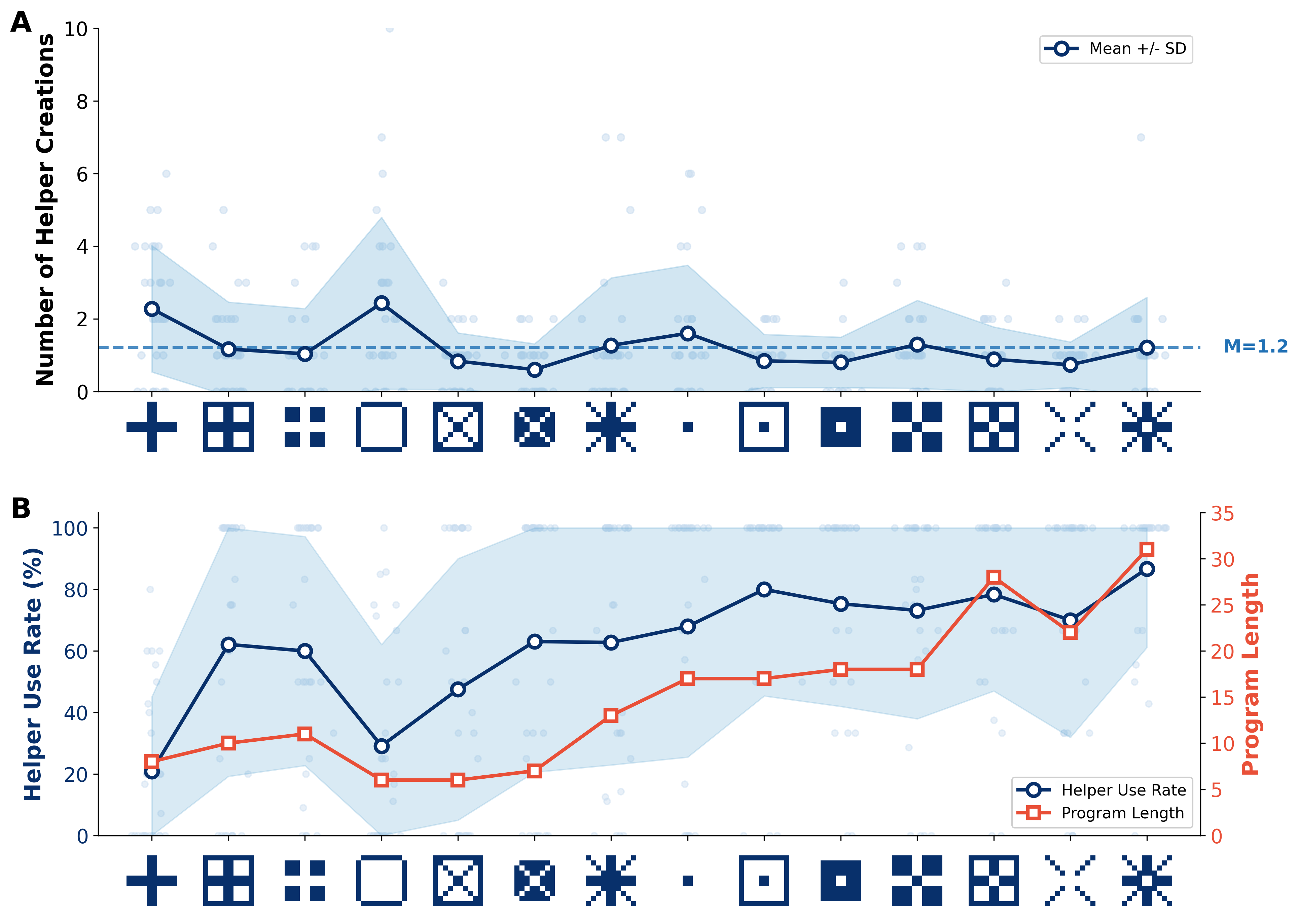}
\caption{Helper usage across 14 target patterns. 
A. Number of helpers created at each trial. B. Proportion of solution steps using saved helpers (blue) overlaid with ground-truth program length (red). }
\label{fig:combined}
\end{figure}

Participants' helper usage became more efficient over time. At the onset of the experiment, participants created the largest number of helpers (P1: $M = 2.28$, $SD = 1.74$; P4: $M = 2.43$, $SD = 2.37$, where \texttt{diagonal} was first introduced; Figure~\ref{fig:combined}A). 
The proportion of solution steps that involved saved helpers increased systematically across trials (Figure~\ref{fig:combined}B): early patterns showed relatively modest helper usage rates (P1: 21\%), whereas later patterns exhibited substantially higher rates (P9: 80\%; P14: 87\%). A linear regression confirmed a significant positive trend in helper usage over trials ($\beta = 3.54\%$ per trial, $r = .79$, $p < .001$).

Consistent with this progression, program length was positively correlated with helper use rate ($r = .77$, $p = .001$), indicating that participants increasingly used helpers when solving longer programs. Together, these findings suggest that participants managed search complexity by expanding the breadth of exploration rather than increasing depth, leveraging reusable abstractions to make search more efficient.

\begin{figure}[t]
\centering
\includegraphics[width=\columnwidth,height=0.5\textheight,keepaspectratio]{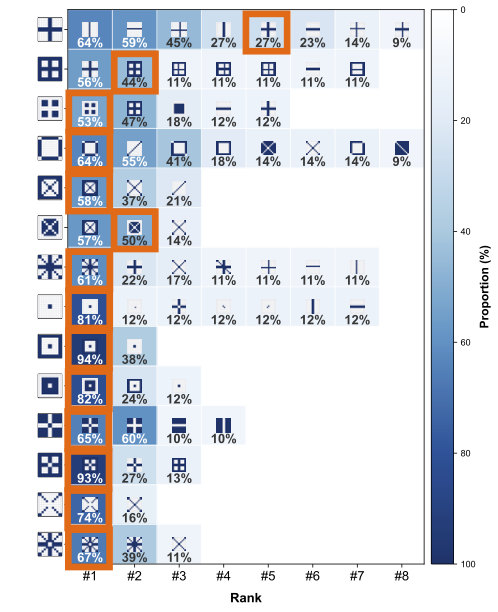 }
\caption{Helpers created by participants.
Each row shows the patterns most frequently saved as helpers for that trial, ranked by popularity. Darker cell color indicates higher  popularity. 
Only patterns saved by more than one participant are shown. Orange borders are for helpers saved by \textit{Library} models.}
\label{fig:distribution}
\end{figure}

In addition to increased helper usage, participants also converged on helper creation strategies.
In early trials (P1--P7), the most common helper was saved by 53--64\% of helper-creating participants. This agreement increased drastically in later trials: P8 (81\%), P9 (94\%), P10 (82\%), and P12 (93\%), 
where nearly all 
participants saved the same patterns (Figure~\ref{fig:distribution}).
This convergence matches the heuristic used in our Library models: adding the final solution as a new primitive. 
Among participants who created helpers, the proportion who saved the target pattern itself increased over time: 50.8\% in the first seven trials versus 78.5\% in the last seven trials ($\beta = 3.25\%$ per trial, $r = .73$, $p = .003$). 


\paragraph{Nodes expanded, but not program length, predicts participants' solution time.}

\begin{figure*}[t]
\centering
\includegraphics[width=\textwidth]{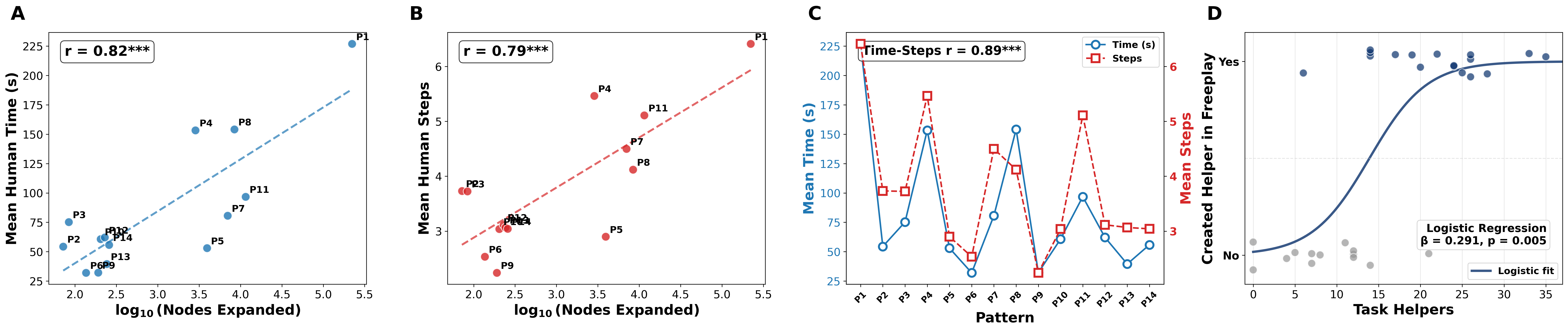}
\caption{Participant performance and model-estimated metrics. 
A. Mean completion time versus $\log_{10}$(nodes expanded). 
Each point for a pattern (P1--P14). Dashed line shows linear fit.
B. Mean number of steps versus $\log_{10}$(nodes expanded). 
C. Mean completion time (blue, left axis) and mean steps (red, right axis) for each pattern. 
D. Number of helpers created during task phase 
predicts helper creation during free play (yes/no). Each point represents one participant, 
with jitter for visibility.}
\label{fig:combined_4}
\end{figure*}

Participants on average spent $84$ seconds ($Mdn = 44.1$s, $SD = 136.1$s) on each puzzle, and completed a pattern with a median of 3.0 steps ($M = 3.8$, $SD = 3.2$). 
These two metrics were strongly correlated across patterns ($r = .89$, $p < .001$) 
(Figure~\ref{fig:combined_4}C).
To examine whether computational models predict human performance, we correlated pattern-level metrics from the \textit{Short+Library} model with human behavioral measures (Figure~\ref{fig:combined_4}). 
The number of nodes expanded strongly predicted both mean solution time ($r = .82$, $p < .001$; Figure~\ref{fig:combined_4}A) and mean number of steps ($r = .79$, $p < .001$; Figure~\ref{fig:combined_4}B). Trial-level mixed-effects models confirmed these relationships: log-transformed nodes expanded significantly predicted steps ($\beta = 0.86$, $z = 6.73$, $p < .0001$) and log-transformed solution time ($\beta = 0.41$, $z = 9.94$, $p < .0001$), accounting for individual differences among participants.

In contrast, while shorter programs were associated with higher success rates($r = -.67$, $p < .01$), program length did not reliably predict solution time or number of steps ($r = -.20$, $p > .05$). This dissociation suggests that what determines human difficulty is not the length of the final program, but the computational cost of discovering it in the compressed space of learned compositional abstractions. 

\paragraph{Helper use shapes free play}
Participants spent an average of 6.32 minutes (beyond the required 5 minutes) in the free-play phase ($SD = 2.15$, $Mdn = 5.66$, range: 5.03--17.12 min). During this time, 27 of 30 participants (90.0\%) submitted at least one pattern to the gallery, yielding a total of 80 creations ($M = 2.67$ per participant, $SD = 1.70$). Of these, 64 patterns (80.0\%) were given custom names by participants, suggesting that people navigated the learned abstract compositional space in a semantically structured way, rather than merely exploring low-level configurations. 

We found that participants often recreated and reused helpers previously built in the task phase, 
suggesting mental sampling from the learned abstract compositional space.
Additionally, 17 of 30 participants (56.7\%) created at least one new helper during free play, with a total of 146 helpers created ($M = 4.87$, $SD = 5.87$). A logistic regression revealed that the number of helpers created during the task phase significantly predicted whether participants created helpers during free play ($\beta = 0.29$, $p = .005$), indicating that engagement with library learning transferred to open-ended exploration (Figure~\ref{fig:combined_4}D). 

\begin{figure}[t]
\centering
\includegraphics[width=\columnwidth, height=0.3\textheight,keepaspectratio]{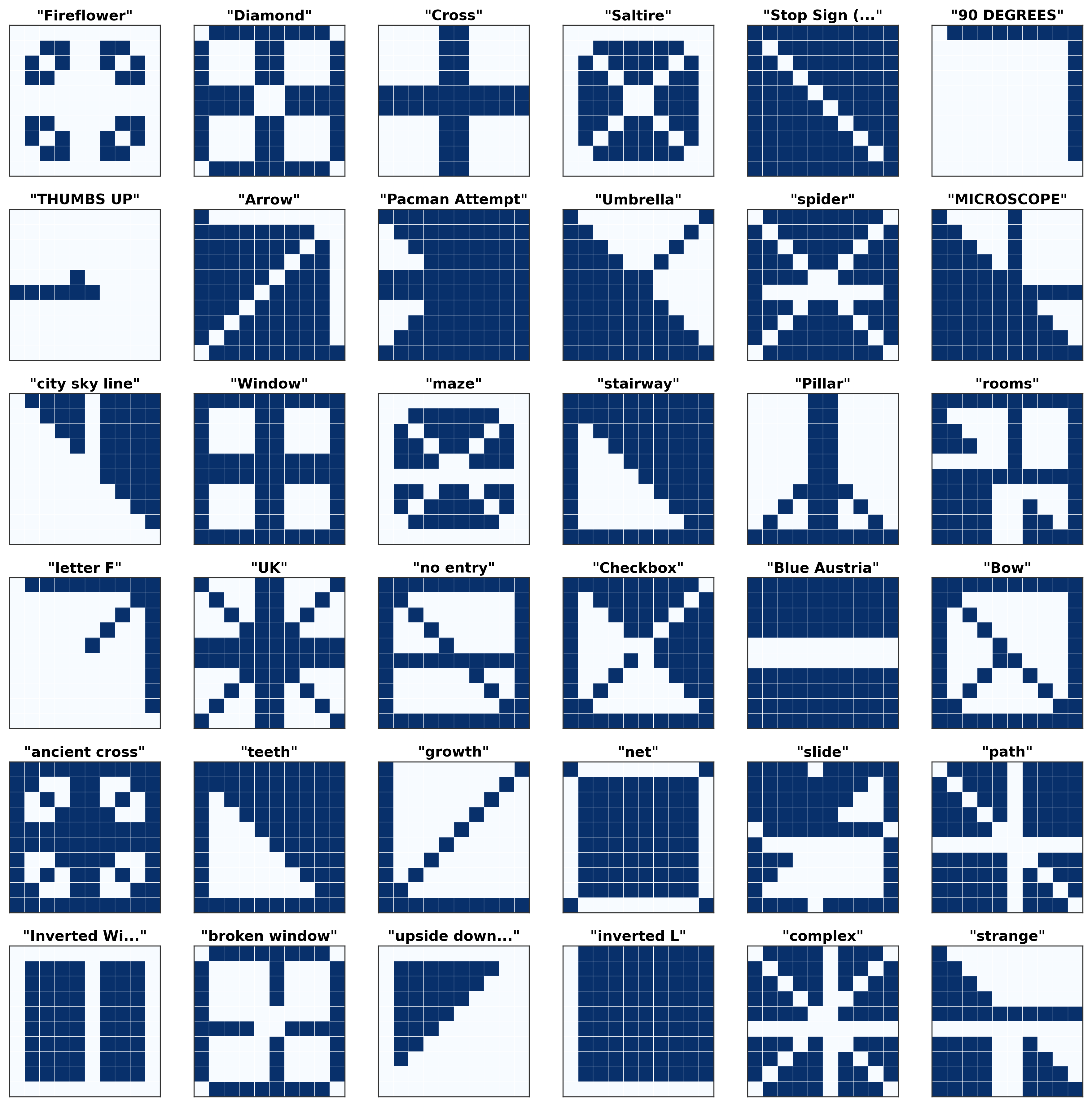}
\caption{Examples of patterns participants created during free play. 
Pattern names were provided by participants.}
\label{fig:freeplay-examples}
\end{figure}

Figure~\ref{fig:freeplay-examples} shows representative patterns created during free play, including regular, symmetric designs (e.g., ``Fireflower''), and semantically-driven compositions (e.g., ``city sky line'', ``THUMBS UP''). Symmetric constructions suggest the use of perceptual priors, consistent with Gestalt theory. Figurative designs suggest that participants exercised top–down semantic control over the learned representations, using them to express prior knowledge about the natural world. 

\section{Discussion}

People naturally create generalizable abstractions 
to make hard tasks easier. They adjust these intermediate representations as they learn about the world around them.
We studied the computational process in this online library learning scenario with 
a novel Pattern Builder Task (PBT) paradigm,
in which participants solve 
complex pattern construction tasks using a small set of geometric primitives, while having the ability to create and reuse intermediate abstractions (“helpers”). This design allows us to observe how reusable structure is constructed, refined, and deployed online as task demands grow.
Our results show that people increasingly rely on helpers when solving visual puzzles, converge on similar abstraction strategies, and engage in online library learning to solve problems that would be otherwise intractable. We formalize the task as program induction with library learning, showing that human solution time and complexity track the model-estimated search complexity (nodes expanded), rather than the length of the shortest primitive program. Finally, in an unconstrained free-play phase, participants continue to reuse and invent helpers, suggesting that abstraction learning is not merely reactive to task difficulty, but functions as a self-sustaining mode of building representations.
Symmetric designs produced in free-play likely reflect Gestalt and compressibility biases, while figurative designs suggest top-down mapping of the learned compositional language onto semantic priors over the natural world.

Participants' helper use was adaptive, given task structure.
Early in learning, participants externalized many intermediate steps, favoring completeness over efficiency. Over time, they became more selective, generating efficient helpers that enabled compressed solutions. Strikingly, this improvement emerged without participants seeing the full task space, and included anticipatory, intermediate structures intended for future reuse.
While our computational model is limited to using complete patterns as new primitives, extending it to capture this partial abstraction could provide a more in-depth account of human problem-solving in online and sequential domains.
Together, the results support the view that human problem solving relies on adaptive, online, and path-dependent learning of compressed library-like abstraction, that restructure the representational vocabulary over time.

Despite these insights, our study focused on individual learners, leaving open questions about how structured abstractions are learned in social contexts, or across generations of learners, where observation, communication, and acculturation could drive the emergence of a common abstraction strategy \citep{acquaviva2022communicating,thomas2024pace,boyce2024interaction,pu2020program,effenberger2021analysis}. Future work could explore how such abstraction libraries evolve in social settings, where helpers can be named, demonstrated, or taught, potentially giving rise to shared representational conventions analogous to cultural tools.




\printbibliography

\end{document}